\pdfoutput=1
\documentclass[11pt]{article}
\usepackage[final]{EMNLP2023}
\usepackage{times}
\usepackage{latexsym}
\usepackage{graphicx}
\usepackage{tabularx}
\usepackage{todonotes}
\usepackage[T1]{fontenc}
\usepackage{adjustbox}
\usepackage{listings}
\usepackage{booktabs}
\usepackage[utf8]{inputenc}

\usepackage{microtype}
\usepackage{graphicx}
\usepackage{amsmath}

\usepackage{inconsolata}
\usepackage{enumitem}
\usepackage{amsfonts}
\usepackage{amsmath}
\usepackage{soul}

\usepackage{comment}
\usepackage[]{todonotes}

\title{HR-Agent: A Task-Oriented Dialogue (TOD) LLM Agent Tailored for HR Applications}

\author{{\bf Weijie Xu}$^{1}$, {\bf Jay Desai}$^{1}$, {\bf Fanyou Wu}$^{1}$, {\bf Josef Valvoda}$^{2}$, {\bf Srinivasan H. Sengamedu}$^{1}$ \\
  $^{1}$Amazon\\
  $^{2}$University of Cambridge \\
  {\tt weijiexu@amazon.com}
}
\begin{document}
\maketitle
\begin{abstract}
Recent LLM (Large Language Models) advancements benefit many fields such as education and finance, but HR has hundreds of repetitive processes, such as access requests, medical claim filing and time-off submissions, which are unaddressed. We relate these tasks to the LLM agent, which has addressed tasks such as writing assisting and customer support. We present HR-Agent, an efficient, confidential, and HR-specific LLM-based task-oriented dialogue system tailored for automating repetitive HR processes such as medical claims and access requests. Since conversation data is not sent to an LLM during inference, it preserves confidentiality required in HR-related tasks. 

\end{abstract}
\section{Introduction}

Recent advances in natural language processing (NLP) have been used for many domains such as Law~\citep{sargeant2024topic}, Finance~\citep{masson-paroubek-2024-evaluating-topic} and Education~\citep{Zhao2021}. However, many HR processes, such as requesting time off, scheduling meetings, submitting tickets for IT issues, or filing medical claims, remain highly inefficient. Automating these processes could save a significant amount of time that would otherwise be spent on repetitive work. This paper investigates how LLM Agent could facilitate such automation. For a LLM Agent to be useful in the HR domain, it must satisfy the following five requirements:

(1) It must have a \textbf{fast} response time. Employees are less likely to use a chatbot if it cannot complete the task quickly \citep{hoxmeier2000system}. Research has shown that user satisfaction decreases as response time increases, and a preferable response time should be less than 2 seconds~\citep{Shneiderma}.

(2) The HR Agent must also be \textbf{extractive}. When using a TOD to file a medical claim, users must be able to trust that the system will accurately retrieve the right number. 

(3) Equally important is the \textbf{versatility} of the system –  it must handle the various HR use cases mentioned above. 

(4) Since employee information is highly sensitive, the TOD itself must be \textbf{confidential}.

(5) Finally, the HR Agent must be \textbf{HR specific} and perform effectively in HR-relevant tasks as suggested by ~\citet{xu-etal-2024-hr}

\begin{figure}

\includegraphics[width=7.8cm]{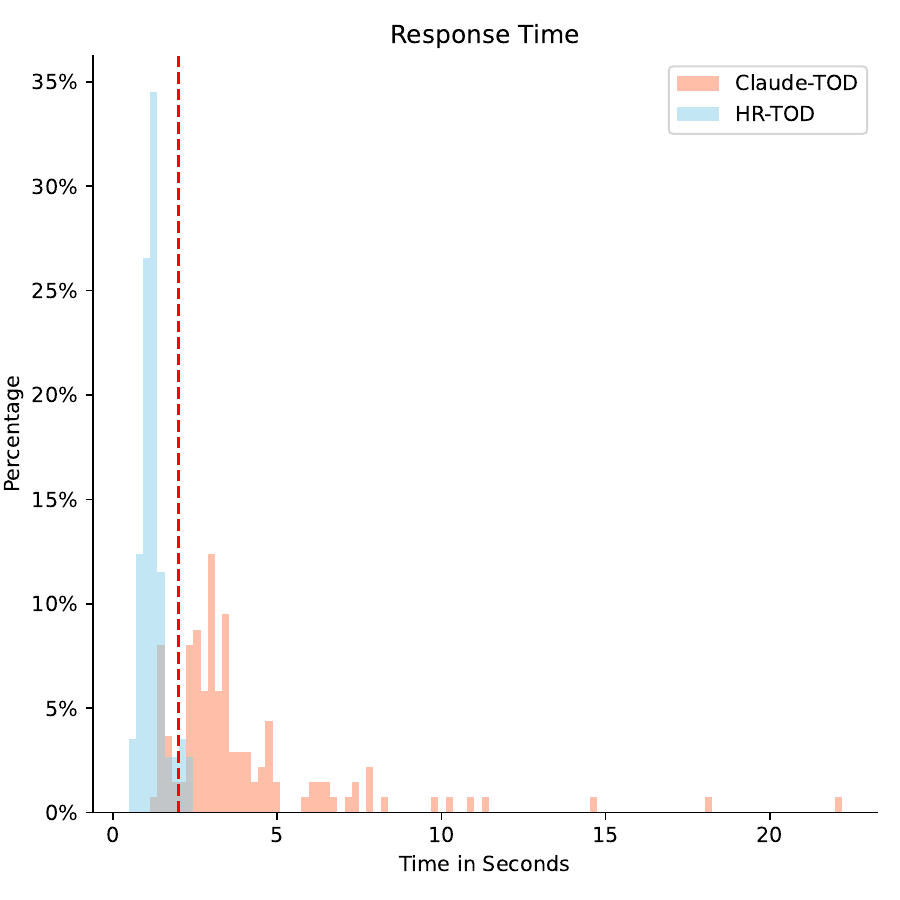}

\centering

\caption{We systematically compared our method to Claude’s approach~\citep{zhang2023sgp} in terms of response time by collecting 40 conversations across four different categories: time off, medical claims, resume creation, and issue ticket handling. The HR-Agent we propose demonstrates significantly faster response times compared to the Claude-based TOD. In fact, our system achieves a response time of less than 2 seconds in 94 percent of cases, while the Claude-based system accomplishes this in only 4 percent of cases. These results highlight the substantial speed advantage of our HR-Agent over the Claude-based solution.\label{time}}

\end{figure}

\begin{figure*}

\includegraphics[width=16cm]{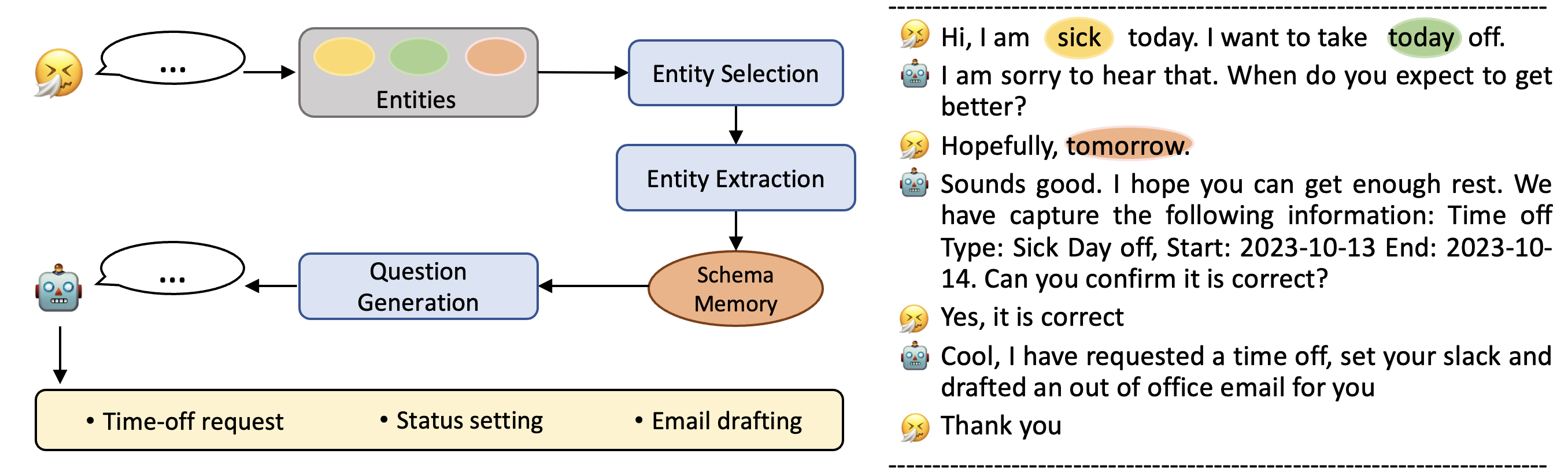}

\centering

\caption{An illustration of the solution. The Entity Selection Model identifies relevant entity. The selected entity is passed to the Entity Extraction model to find the relevant word in the utterance. Based on Schema's memory and previous utterance. The Question Generation model is used to generate the next question. The HR-Agent system then connects to the API to finish the relevant tasks such as drafting email, requesting time off and setting status.\label{systemplot}}

\end{figure*}

To complete the task, these systems rely on Dialogue State Tracking (DST) \cite{rastogi2020towards}, which monitors and predicts user intent and details during conversations. 

DST uses schema-guided techniques, categorized as extractive\citep{rastogi2020scalable, ruan2020fine}, generative~\citep{feng-etal-2021-sequence, tian-etal-2021-amendable}, and LLM-based~\citet{hudevcek2023llms, zhang2023sgp}, to track conversation states with natural language explanations. However, extractive and generative methods are limited by training data and poor transfer learning, hindering HR-specific and versatile applications. They therefore do not satisfy points 3 and 5 above. LLM-based approaches are slow and may lack confidentiality, they therefore do not satisfy requirements 1, 2 and 4. In response, we propose an HR-specific LLM Agent (HR-Agent). It is fast in response time, extractive, versatile, confidential, and HR-specific. We summarize our contributions:\looseness-1

\begin{itemize}[leftmargin=*]

\item Despite being smaller and faster in speed, as illustrated in Figure~\ref{time}, modules trained on synthetic datasets achieve superior performance compared to larger language models.

\item Based on these modules, we designed an HR-Agent system that significantly improves HR process efficiency.

\end{itemize}

\section{Related Work}
Schema-Guided Dialogue (SGD) \citep{rastogi2020scalable} is a dialogue dataset with evolving ontologies, introducing new test set slots and services, emphasizing DST performance and zero-shot generalization. SGD-X \citep{Lee_2022} expands on SGD, presenting five additional schema styles. MultiWOZ~\citep{budzianowski2020multiwoz} features human-human dialogues using a stable ontology. HR-MultiWOZ~\citep{xu-etal-2024-hr} is aligned with HR related tasks, but it is too small to be used for training. \textit{Collecting data for these datasets is labor-intensive and costly}. 

SGD-baseline~\cite{rastogi2020scalable}, SGP-DST~\cite{ruan2020fine}, and DS-DST~\cite{zhang-etal-2020-find} encode utterances and slot schemas jointly to predict relative slots. Multi-Task BERT adopts slot carryover mechanisms, encoding only the preceding system utterance and the current utterance. LUNA~\cite{wang-etal-2022-luna} separately encodes dialogue history, slots, and slot values, learning to predict the correct utterance to condition slot value predictions. However, \textit{these methods lack HR specificity and versatility.} Seq2Seq-DU~\cite{feng-etal-2021-sequence} and AG-DST~\cite{tian-etal-2021-amendable} derive states differently, while DaP~\cite{lee-etal-2021-dialogue} offers two versions, with the latter being slower. D3ST~\cite{zhao-etal-2021-effective-sequence} deciphers the entire dialogue state at once. Although these generative methods, especially when using T5 models, achieve better Joint Goal Accuracy (JGA) in schema-guided dialogue, they suffer from \textit{slow response times} because of extensive input.

Despite recent research efforts~\citep{li2023guiding, hudevcek2023llms, zhang2023sgp}, LLM based TOD’s performance metrics like BLEU scores and success rates remain low, even with models like Alpaca-LoRA-7B~\citep{taori2023alpaca} and ChatGPT. LLMs are also non-extractive and suffer from \textit{high inference costs and latency issues~\cite{yang2023harnessing}}, making real-world TOD system deployment challenging. For instance, using the GPT-4 8K context model via OpenAI’s API costs \$0.03 for every 1K input tokens and \$0.06 for every 1K output tokens.

Gan et al. (2024) introduced a framework using LLM agents for automated resume screening, improving recruitment efficiency through automated extraction and analysis. It achieved high accuracy and F1 scores in sentence classification and effectively protected privacy by excluding personal information while grading and summarizing resumes. However, unlike HR-Agent, it does not specifically address the variety of repetitive HR tasks, such as access requests, medical claim filing, and time-off submissions \citep{gan2024application}.

Enhancing LLM-based task-oriented dialogue systems by incorporating goal-oriented rewards and reinforcement learning techniques was explored by Singh \citep{singh2023exploring}. By aligning dialogue responses with predefined goals, the model improves contextual appropriateness and task-specific outputs. This approach uses MultiWOZ datasets to measure success and refines dialogue policy models through reinforcement learning. Unlike HR-Agent, Singh’s work does not focus on the confidentiality and specific needs of HR-related tasks.

\section{Methods}
We first used an HR task-specific schema from ~\citep{xu-etal-2024-hr}. We then use the question and answer from that dataset to be our training set. We care less about out of utterance response. Subsequently, we format the question and the utterance as an entity extraction task and select the most relevant entity from the utterance. In practice, once we have gathered sufficient schema information, we employ a few APIs to use this structured information for tasks such as drafting emails, creating tickets, and answering questions as illustrated in Figure~\ref{systemplot}. Our designed HR-Agent system can be used for tasks such as requesting leaves, inquiring about benefits, applying for internal jobs, navigating the onboarding process, requesting training, reporting workplace issues, participating in surveys, and engaging with HR initiatives. It simplifies benefit enrollment, goal setting, safety guidelines, and compliance training. It is confidential because the external LLM is not used for any inference. We use synthetic data from \citet{xu-etal-2024-hr} to complete the experiment.

\begin{table*}[ht]
\begin{center}
\begin{tabular}{lllllll}
\hline
          & FlanT5 XL & Falcon & MPT    & FlanT5-Clean & FlanT5-Raw  & Claude V3 \\ \hline
Size      & 3B         & 7B     & 7B     & 220 M    & 220 M&        \\ 
Precision & 0.581      & 0.948  &\textbf{ 0.950 } & 0.910    & 0.777 & 0.392      \\ 
Recall    & 0.881      & 0.753  & 0.754  & 0.832    &0.663 &\textbf{ 0.852}       \\ 
F1 Score  & 0.663      & 0.826  & 0.828 & \textbf{0.856}   &0.703  & 0.519       \\ \hline
Response time & 1.202  &1.085  & 0.6293  & \textbf{0.366 }    & 0.384 & 1.163     \\ \hline
\end{tabular}
\end{center}
\caption{The performance for entity selection. From left to right, we document the performance of FlanT5 XL, Falcon 7B, MPT 7B, FlanT5 base trained on filtered data, FlanT5 base trained on unfiltered data, ClaudeV3.}
\end{table*}

\begin{table*}[ht]
\begin{center}
\begin{tabular}{llllllll}
        & Flan T5 XL & Falcon & MPT   & Deberta & Roberta & Flan T5 Trained & Claude V3 \\\hline
Size    & 3B         & 7B     & 7B    & 135M     &  125M       & 220 M           &  \\

RougeL & 0.786      & 0.112  & 0.292 & 0.729   & 0.767   & \textbf{0.818 }          & 0.793\\\hline
Response time & 0.394       & 1.083  & 0.569 & 0.191  & \textbf{0.088}   & 0.110            & 1.22\\\hline
\end{tabular}
\caption{The performance for entity extraction. From left to right, we document the performance of FlanT5 XL, Falcon 7B, MPT 7B, Debetra, Roberta, FlanT5 base trained on filtered data, and ClaudeV3. }
\end{center}
\end{table*}

\subsection{Baseline Methods}
\textbf{Falcon}~\citep{falcon40b}, a Large Language Model (LLM), is designed for task-oriented dialogue systems. It focuses on optimizing efficiency and utility, aiming to provide fast, accurate responses in various conversational scenarios. 
\textbf{MPT}~\citep{MosaicML2023Introducing} is a transformer trained from scratch on 1T tokens of text and code. It is an open source, available for commercial use, and matches the quality of LLaMA-7B 
\textbf{Deberta}~\citep{he2021deberta} Deberta enhances the BERT model with a disentangled attention mechanism for more interpretable attention scores and uses relative position encoding to boost performance in extractive task. 
\textbf{Roberta} RoBERTa~\citep{liu2019roberta} is a transformer model pretrained on English data using the Masked Language Modeling technique, masking and predicting 15 percent of input words for bidirectional learning. We use both models finetuned on SQuAD2.0 which performs well on extractive QA. 
\textbf{FlanT5} we use the same setup as~\citep{lin2021leveraging} where we use DDP to train, set up validation loss for early stopping and set max epochs equal to 20.

\subsection{Entity Selection}
For entity selection, we need to select the relevant entities that could be answered by utterance. We choose FlanT5 as our base model because it contains smaller model size and is explored most in Schema Guided Dialogue literature. Since the entity itself is not informative, we choose to generate questions about the entity because datasets that are used to train Flan are mostly contain questions. Next, we assess 5 models using the test set. We give the same example for each template and average the result of 10 templates for a pretrained model, such as MPT and Falcon. FlanT5 is trained 5 times and the evaluation result is averaged. The model is trained and evaluated using p3.8xlarge. We do not use an accelerator or ONNX for inference to get a fair comparison of the response time. We benchmark Precision, Recall, F1 Score and Response Time. 

Our ideal solution should have high recall but should not output too many entities, which could drastically increase the response time of the TOD system. Decoder-based models perform well in F1 Score but are low on recall. The fine-tuned version of FlanT5 achieves the best F1 score while also achieving the second-best recall. Claude V3 achieves the best recall but tends to select most of the entities and is slow in response. Without data cleaning, FlanT5 tends to choose many first choices, which leads to low recall. We choose FlanT5 trained on filtered data as our entity selection model because it is \textbf{fast} and ia \textbf{confidential}. Since the training set is \textbf{versatile and HR specific}, the trained model also inherits these advantages. 

\subsection{Entity Extraction}
Except Falcon and MPT, we also use Deberta and Roberta for the entity extraction benchmark. We also fine-tuned a small FlanT5 model on synthetic dataset(20K). We compare Rouge 1, Rouge L and Response time for each model. Our goal is to select the model that is extractive and has low response time.

Our ideal solution has a low response time and a high Rouge1 score. Claude does not achieve a good Rouge1 score and takes much longer to train. It fails to use words from the utterance. Based on our selected models, we find that extractive models such as Roberta and Deberta perform well in the extraction task, while decoder-only models such as MPT and Falcon perform poorly. Flan T5 performed the best. Flan T5, trained on a filtered synthetic dataset, further improves the Rouge1 performance by another 5 percent since it is more \textbf{extractive}. The trained model is also very \textbf{fast} compared to larger models and is comparable to Roberta-based models. Thus, we choose trained FlanT5 as our model for entity extraction.

\subsection{Question Generation} 
Infusing more empathy into the responses is paramount, as it fosters a more human-centric and relatable interaction experience. To achieve that, we use the utterance and the next question to ask as input. We provided this input to Claude-V3 and asked Claude-V3 to rewrite a concise response with empathy, as illustrated in Table~\ref{empathy}. To evaluate the performance of Claude3-V3 on rephrasing response with empathy, we want to benchmark against the response generated in the HR-Multiwoz dataset. To achieve that, we first identify response that may reflect negative feelings from the user. We use DistillBert based on a case model finetuned on dataset SST-2 as our sentiment classifier. Since finetuned DistillBert is uncalibrated. Only sentences with a score above 0.998 and classified as negative are selected. We have collected 638 responses from HR-Multiwoz. We then manually rewrite those questions to basic question and leverage Claude3-V3 to rewrite it. Basic question means it only contain question itself without any additional information.

To evaluate the effectiveness and human preference for the responses generated by our HR-Agent, we conducted a preference study involving human labelers using Amazon SageMaker Ground Truth (GT). We compared responses generated by HR-Agent (denoted as Response A) with those from the HR-MultiWOZ dataset (denoted as Response B).

\begin{figure}[ht]
\centering
\includegraphics[width=0.5\textwidth]{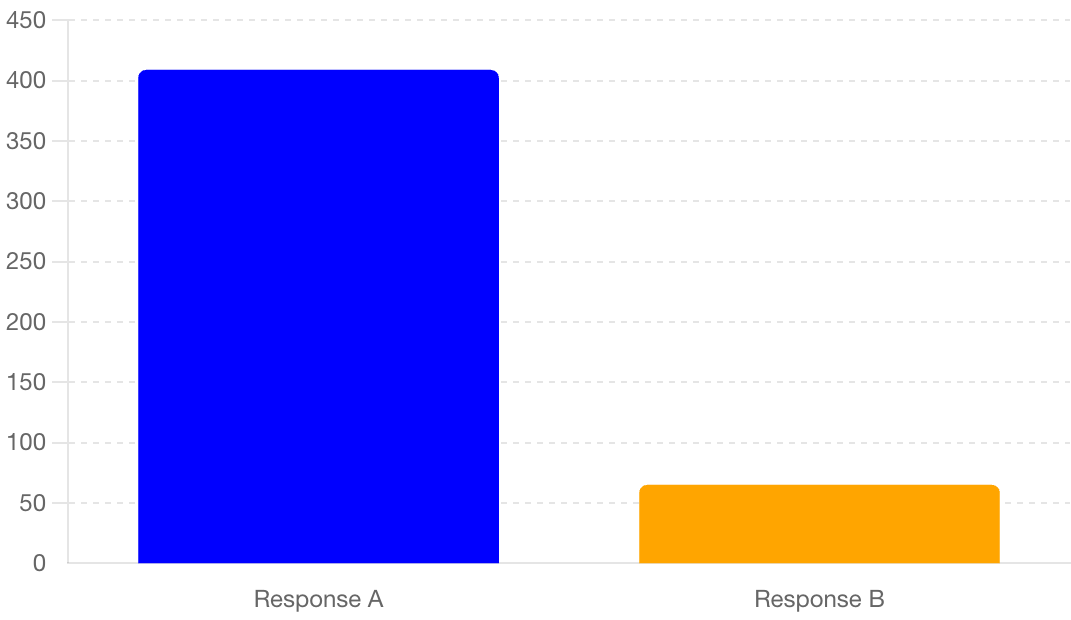}
\caption{Labeler preferences for responses from HR-Agent (A) and HR-MultiWOZ (B).}
\label{fig:labeler-preferences}
\end{figure}

Labelers were presented with dialogue scenarios and asked to choose which response they preferred. The user interface used for this evaluation is shown in Figure \ref{human_preference_ui}. Each response was evaluated by three labelers, and the cost per label was \$0.0012. The results of this study are illustrated in Figure \ref{fig:labeler-preferences}. 

Our study revealed a significant preference for responses generated by HR-Agent. Out of the total responses evaluated, Response A was preferred 409 times, while Response B was preferred only 65 times. This strong preference for Response A suggests that the HR-Agent’s responses were more aligned with the labelers’ expectations and needs.

The HR-Agent’s use of empathy in responses contributed to its preference. For instance, when an employee inquired about scheduling a medical appointment, the HR-Agent’s response, “I understand that medical emergencies can be stressful. Please share the incident details so we can assist you,” was perceived as more empathetic and supportive compared to a more straightforward response from the HR-MultiWOZ dataset.

These findings underscore the value of tailoring dialogue systems to specific domains and incorporating elements such as empathy to improve user satisfaction and engagement. 

\subsection{System Evaluation}

To demonstrate that our designed system performs best in HR related tasks. We use HR-Multiwoz as our evaluation set. We compared our method with TransferQA~\citep{lin-etal-2021-zero} and SGP-TOD~\citep{zhang2023sgp}. ZST adapts a sequence-to-sequence model (T5) pre-trained on large QA datasets to track dialogue states by formulating slot-filling tasks as QA problems. We replace T5 by Deberta and Roberta, suggested by HR Multiwoz benchmark leaderboard. SGP-TOD uses prompt engineer to guide the dialogue and extract the relevant fields. 

Joint Goal Accuracy (JGA) and Average Goal Accuracy (AGA) are used to evaluate our models and baselines. For JGA, the model outputs are only counted as correct when all the predicted values exactly match the oracle values. AGA is the average accuracy of the active slots in each turn. 

As you can see in Table~\ref{fig:all}, our proposed method achieves the best performance in HR-Multiwoz dataset in terms of JGA and AGA compared to the state-of-the-art methods. This means that the HR Agent works well in HR related SGD tasks.

\subsection{Prompt}
Prompt we used in the experiment for MPT, Falcon and FlanT5 as you can see in Table~\ref{tab:sentences1} and Table~\ref{tab:sentences2}.

Prompt for Claude-V3 is  "The answer is very short and always less than 2 words. Put the answer in <answer></answer>  XML tags." for entity extraction and "The answer always contains 2 to 5 choices. Put the answer in <answer></answer>  XML tags." for entity selection. We use max tokens to sample be 10, temperature be 0.2 and stop sequence be </answer>

\begin{table}[ht]

\begin{center}
\begin{tabular}{lll}
\hline
          & JGA & AGA  \\ \hline
TrasnferQA Roberta      & 8.65       & 26.62       \\ 
TrasnferQA Deberta & 18.89 & 55.61      \\ 
SGP-TOD    &10.74   &54.77       \\ 
HR-Agent  & \textbf{20.47}  &  \textbf{55.38}  \\ \hline
\end{tabular}

\end{center}
\caption{The overall performance of HR Agent on HR-Multiwoz. It achieves better performance in JGA as well as AZA.}
\label{fig:all}
\end{table}

\subsection{Other considerations} 
An essential part of this process involves robust fact-checking to validate generated answers, enhancing reliability and credibility. For instance, when identifying a medical provider, the system confirms their details in the database before processing a medical claim. Users can also track the schema throughout the process, ensuring accountability and trustworthiness. Our dataset connects with existing databases to retrieve relevant information and only asks when necessary. After collecting data, users confirm the schema and the task, and we connect to relevant APIs to complete the task. Before putting information into the system, it is important to make information accurate. For example, the time off day collected is today is not informative for the system to track the exact day. To resolve these inaccuracies, we call Claude to auto complete the inaccurate information. For example, the Claude can change tomorrow to November 1st and 98121 to Seattle WA. This can help the system capture the accurate information. We then confirm it with the user before calling APIs to use the structure information. The system can hugely reduce user’s inputs to get required information.

\section{Conclusion}
In improving the efficiency of corporate employees, we introduce HR-Agent, a rapid, extractive, and confidential dialogue system tailored to HR needs. Our work encompasses several key contributions: identifying challenges in adapting TOD to HR contexts, devising a swift and domain-specific data generation approach, demonstrating the superior performance of smaller, faster-trained modules over larger models, and ultimately delivering HR-Agent as a solution that markedly enhances HR process efficiency.

\section{Ethical Statement}
This system is a prototype and has not been deployed in production. 

The deployment of AI applications in the HR space necessitates careful consideration of ethical issues related to safety, privacy, and bias. There is a possibility that, in attempting to assist, AI may cause more harm than benefit. In response, in collaboration with user experience researcher, security reviewer and HR professionals, we have suggested the following steps for developers who plan to use the HR Agent in order to minimize the risks of harm.

Informed Consent from Users: In the pilot phase, informed consent was obtained from employees using the service. They were made aware that they would be interacting with an AI-based chatbot designed to expedite task completion. It was also communicated that some extracted information might be inaccurate, and users had to verify the correctness of this information before utilizing it for subsequent tasks. To facilitate this, HR Agent should present the collected information at the end of each conversation and request employee confirmation of its accuracy. Developers should ask the employees to participate in a survey after using HR Agent to understand the effectiveness and helpfulness of HR Agent. They should make it clear that HR Agent is used to fill the relevant task and not for anything else. They should provide relevant HR Business Partner contact information as headers to employees when they interact with HR Agent.

Guardrail:HR agent should pose clarifying questions when unable to extract relevant entities. The conversation should terminate if clarifying questions for the same entity are repeated more than three times. The developers can use sentiment analysis model to monitor the sentiment of employees every four user responses, and the conversation is ended if the negative sentiment score exceeds 0.5. Developers should provide task relevant internal wiki when they end the conversations. For some applications, guardrails are also in place for ranges of money and time that exceed specific thresholds. It's important to note that HR Agent has limited applications at present, and they should update these guardrails accordingly.

Privacy: In the system, information provided by the employees should be kept confidential in a record separate from the employee’s general personnel file. Employees have the right to use this record or input this record in other systems or give developers rights to use this record for analysis. This policy follows the Americans with Disabilities Act (ADA and the Genetic Information Nondiscrimination Act (GINA). Please note that all the models in the system should be trained on synthetic data. Developers should not use any real employee data to train the model. Developers also have to make sure the data in the system is in compliance with rigorous internal infoSec policies and standards. For example, security testing includes examining application logs to detect any data leakage into logs.

Negative Examples/Potential Bias: To mitigate potential biases in generative models, developers should employ an extractive approach. None the less, the effectiveness of extractions could vary with the employee's language fluency. This variation could potentially lead to inefficiencies in the Task-Oriented Dialogue (TOD) system for non-native English speakers. Additionally, the system is prone to errors when extracting multiple entities of the same type, such as time and money from one response. Efforts are underway to understand and address these issues.

Developers should do threat modeling, security testing, penetration test assessment of the system. 

\section{Limitations and Risks}
\textbf{Limitations} We have not explore other models for training such as Deberta or LLama2. We have not evaluated the model on real data. 

\textbf{Risks} Because of some privacy concerns, we have not discussed the details of the whole architecture. The risk of deploying this architecture is that the performance may drift a bit on real data since test set is synthetic.

\section{Future Work}
Looking ahead, potential advancements include the capability to generate resumes and emails, interface with various APIs, proficiently answer queries, and identify pertinent tickets, further enhancing the utility and efficiency of the system. To enhance the performance and capabilities of the model, it is crucial to amass a substantial dataset for the fine-tuning of a more extensive model, enabling it to execute multiple tasks concurrently. Essential prerequisites for this advancement encompass the ability to handle longer sequences and the provision of clear, upfront instructions for entity extraction and task execution. Furthermore, illustrating the connections to other agents is vital, offering a comprehensive and interconnected approach to task management and execution, thereby bolstering the overall efficiency and effectiveness of the model. We could leverage topic modeling~\citep{xu-etal-2023-vontss, xu-etal-2023-detime, xu2024kdstmneuralsemisupervisedtopic,guo2024jadsframeworkselfsupervisedjoint} to better understand and segment use cases. We could leverage differential private synthetic data generation mechanism~\citep{madl2023approximateadaptanonymize3a,xu2023ffpdgfastfairprivate} to avoid potential privacy issues. We can also apply it in hiring process ~\citep{10.1145/3674847,fang2024large}.

\bibliography{anthology,custom}
\bibliographystyle{acl_natbib}

\appendix

\section{Synthetic Example}
\label{claude_schema}
We first try to generate data that is similar to these datasets where we have the conversation and schema generated by the conversation. We can use two claude agent where one acts as an employee and another acts as a TOD system. We then use the third claude agent to track the schema. To generate the conversation, the approach needs multiple api call per sample which is costly. 

There are also many other problems. After a few rounds of generation, the result becomes repetitive even with dramatically different prompts.   The language model tends to follow the most probable or common paths given their training data and their priors, and they lack the creativity or the incentive to explore alternative. The conversation can also be irrelevant. Thus, the system trained on these data is less likely to be extractive, versatile and HR specific. 

\begin{table*}[ht]
\centering
\begin{tabularx}{\textwidth}{|lX|}
\hline
\textbf{Speaker} & \textbf{Dialogue and Schema} \\
\hline
Employee & I'd like to take next Thursday and Friday off. \\
\hline
Chatbot & Okay, which dates would you like off? $\rightarrow$ \text{Ask repetitive questions} \\
\hline
Schema & timeOffStartDate: ? $\rightarrow$ \text{Not able to capture the right information} \\
      & timeOffEndDate: ? \\
\hline
Employee & Next Thursday the 19th and Friday the 20th. \\
\hline
Chatbot & Got it. Can you confirm the reason for the time off? \\
\hline
Schema & timeOffDate: October $19$-$20$, $2023$ $\rightarrow$ \text{Suddenly change the schema and not extractive} \\
\hline
Employee & It's for a doctor's appointment. My name is John Smith. \\
\hline
Chatbot & Okay, I have a request from John Smith for October 19th-20th for a doctor's appointment. \\
\hline
Schema & timeOffDate: October $19$-$20$, $2023$ \\
      & reason: Doctor's appointment \\
      & name: John Smith \\
\hline
\end{tabularx}
\caption{This is a generated example by Claude. The schema keeps changing. The schema fails to capture correct information. The chatbot also asked the information that has been mentioned by user already. The chatbot is also not extractive}.

\end{table*}

\section{Designed Prompt}

Here is our designed prompt where \textit{schema}, \textit{number1} and \textit{number2} change every time. We use output1 and output2 to capture answers for two tasks. We use Claude-V3 model. We use max toke sample equal to 4096. We keep temperature equal 1, top k equal 1 and top p equal 0.6 to have good output length, capture multiple samples at a time and have enough diversity. 

\begin{table*}
\small
\centering
\label{prompt}
\begin{tabular}{|p{0.95\textwidth}|}
\hline
\textbf{Human:} You are asked to come up with a set of 20 diverse scenarios. The input is a user response and a list of questions that the user response could answer about the user. The instruction is to select the right set of questions that could be answered by user input and give an answer for each selected question. \\
\hline
\textbf{Requirements:}
\begin{enumerate}[label=\arabic*.]
\item Try not to repeat the verb or cases for each input to maximize diversity.
\item The language used for the user response should be diverse.
\item The type of input should be diverse. The user response should include diverse types of tasks like \textit{{schema}}.
\item You should generate an appropriate List of Questions for the input. It should involve realistic data and should not contain simple placeholders.
\item The list of questions should start with a choice name such as a, b, c, ...
\item Output1 should be the choice that could be answered by the user response.
\item Output1 should be all correct choice names such as a, b, c.
\item The list of questions should contain \textit{{number1}} questions.
\item Output1 contains \textit{{number2}} choices.
\item Every question has an equal chance to be the correct answer. The answer should not always contain 'a'.
\item Output2 should be the correct answer for each question.
\item The answer from Output2 is extracted from the User Response.
\item The answer of Output2 cannot be 'yes' or 'no'.
\item The scenario is to help corporate extract information from their employees.
\item Try not to ask similar questions to maximize diversity.
\item Output2 does not contain choice symbols like a, b, c.
\item The answer from Output2 can only be extracted from the User Response.
\item Output2 uses a comma to separate the answers.
\item Two empty lines between each case.
\end{enumerate} \\
\hline
\textbf{Example:}

\textit{User Response: I am taking next Monday off as a vacation day.}

\textbf{List of Questions:}
\begin{enumerate}
\item[a.] When is the requested time off?
\item[b.] What action does the user want the recipient to take?
\item[c.] What process has the user completed?
\item[d.] What type of time off is being requested?
\end{enumerate}
\textbf{Output1:} a, d \\
\textbf{Output2:} next Monday, vacation day \\
\hline
\end{tabular}
\caption{This is an example prompt for claude for generation\label{prompt_claude}}

\end{table*}

\section{Output Example}
\label{other_option}
Here is an example of the output using Claude V3:

\begin{table*}[ht]
\centering
\label{example}
\begin{tabular}{|p{0.95\textwidth}|}
\hline
\textbf{User Response:} I would like to schedule a doctor's appointment for next Tuesday at 2pm to get a physical exam.

\textbf{List of Questions:}
\begin{enumerate}[label=\alph*.]
\item What type of appointment does the user want to schedule?
\item When does the user want to schedule the appointment?
\item What time does the user want the appointment?
\item What is the purpose of the appointment?
\item What action does the user want the recipient to take?
\item On what date is the user requesting the appointment?
\item Does the user provide the date for the requested appointment?
\item Does the user provide the time for the requested appointment?
\end{enumerate}
\textbf{Output1:} a, b, c, d \\
\textbf{Output2:} schedule, next Tuesday, 2pm, physical exam

\end{tabular}
\caption{The example perfectly follow our designed prompts using Claude. \label{example_calude} }
\end{table*}

We have also tried cheaper option such as Jurassic-2 but we fail to make them follow instruction.

\begin{table*}[ht]
\centering

\begin{tabular}{|p{0.95\textwidth}|}
\hline
Here is an example of Jurassic-2
\textbf{User Response:} "I would like to inquire about my employee benefits."

\textbf{List of Questions:}
\begin{enumerate}
\item What type of benefits would you like to inquire about? (e.g., health insurance, retirement plans, etc.)
\item Are you currently employed by the company?
\item How long have you been with the company?
\item Are you eligible for benefits? (e.g., full-time employee, part-time employee, etc.)
\item Are you the primary beneficiary of any benefits? (e.g., spouse, children, etc.)
\item Do you have any specific questions or concerns about your benefits?
\item How would you prefer to receive your benefits information? (e.g., email, letter, etc.)
\end{enumerate}

\textbf{Output 1:}
\begin{enumerate}
\item[a)] health insurance
\item[b)] retirement plans
\item[c)] life insurance
\item[d)] dental insurance
\item[e)] disability insurance
\item[f)] vacation and sick leave
\end{enumerate}

\textbf{Output 2:}
\begin{enumerate}
\item[1.] health insurance
\item[2.] yes
\item[3.] full-time employee
\item[4.] yes
\item[5.] yes
\item[6.] no
\item[7.] email
\end{enumerate}

\end{tabular}
\caption{Example generated by Jurassic-2}
\end{table*}

\section{Experiment Setup Appendix}
\label{setup}

Various tables related to Experiment Setup are shown in this section.

\begin{table}[ht]
\centering
\begin{tabular}{|c|l|}
\hline
\textbf{No.} & \textbf{Sentence} \\
\hline
1 & Could you retrieve the answer to the Question from the Text? \\
\hline
2 & Can you pull out the response to the Question within the Text? \\
\hline
3 & Would you mind extracting the reply to the Question from the Text? \\
\hline
4 & I'd like you to get the answer to the Question from the Text. \\
\hline
5 & Could you find the solution to the Question in the Text? \\
\hline
6 & Please identify the answer to the Question in the Text. \\
\hline
7 & Can you locate the response to the Question from the Text? \\
\hline
8 & I'd appreciate if you could extract the answer to the Question from the Text. \\
\hline
9 & Would it be possible to get the reply to the Question from the Text? \\
\hline
10 & Please search for the answer to the Question within the Text. \\
\hline
\end{tabular}
\caption{Table format prompt for entity extraction}
\label{tab:sentences1}
\end{table}

\begin{table}[ht]
\centering
\begin{tabular}{|c|l|}
\hline
\textbf{No.} & \textbf{Sentence} \\
\hline
1 & Could you identify the appropriate question that Text can answer? \\
\hline
2 & Please find the correct question for which the Text provides an answer.  \\
\hline
3 & Can you determine the suitable question that can be resolved using the Text? \\
\hline
4 & I'd like you to pinpoint the right question that the Text can address.  \\
\hline
5 & Please locate the question that aligns with the Text's answer.\\
\hline
6 & Could you discern the fitting question that the Text can respond to?  \\
\hline
7 & I'd appreciate if you could determine the exact question that can be answered using the Text.  \\
\hline
8 & Can you select the question that the Text can satisfactorily answer? \\
\hline
9 & Would it be possible to identify the question that matches the Text's answer?  \\
\hline
10 & Please deduce the right question that aligns with the Text's response. \\
\hline
\end{tabular}
\caption{Table format prompt for entity selection}
\label{tab:sentences2}
\end{table}

\section{Data Validation}
\label{validation}
\subsection{LLM use for Validation data}
We use the following string to put into Vicuna and FlanT5 XXL to validate the data: Question: ${question}$  Text: ${text}$   Answer: ${answer}$ Does tha Answer answer the Question based on Text? The answer could be yes or no" We only select data with answer contains "yes" For Claude, we input 50 in a group and ask model to find line number with the answer equal no. We save inference cost this way. 

\subsection{Mechanical Turk}
We give 0.012 dollar per human labeller per task. We do not enable automated data labeling.

\begin{figure*}[ht]
\centering
\includegraphics[width=16cm]{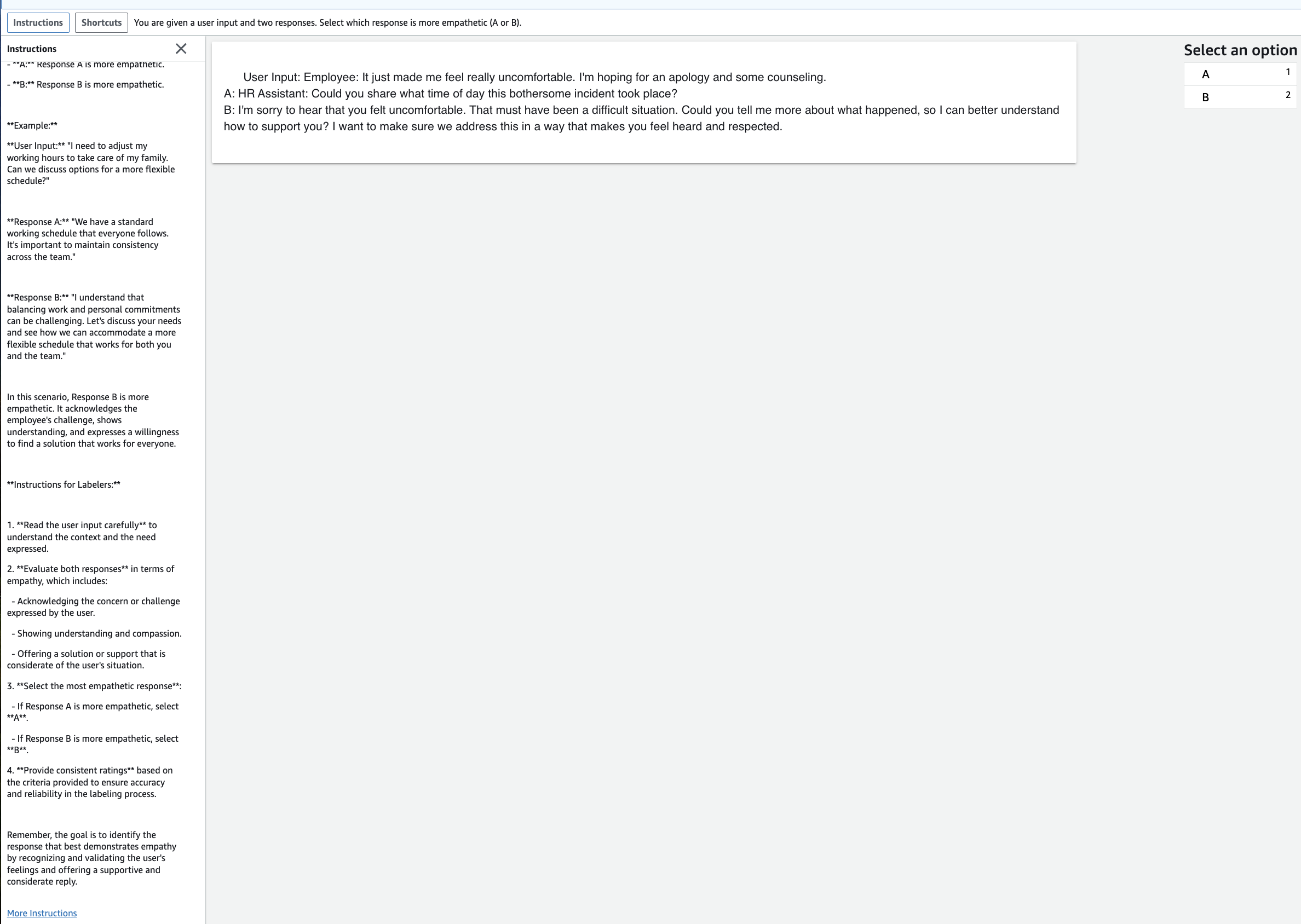}
\caption{User interface used for the human preference study.}
\label{human_preference_ui}
\end{figure*}

\begin{figure*}
\includegraphics[width=16cm]{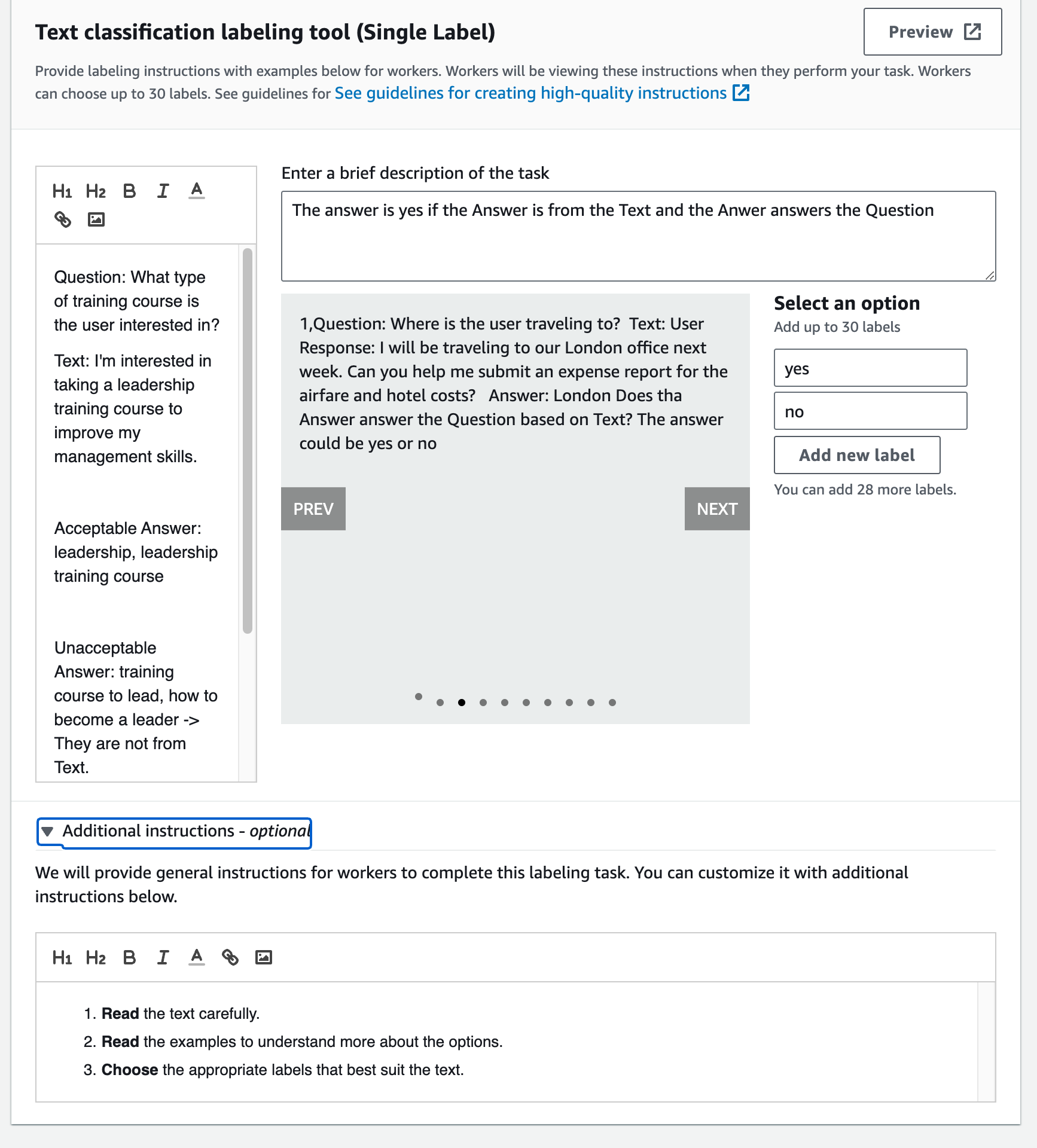}
\centering

\caption{MTurk Questions and selected examples}
\end{figure*}

\clearpage

\section{Question Generation Data}
The following gives an example to illustrate the synthetic sample. 
\begin{table}[htbp]
\centering
\caption{User Input, Next Question, and Empathetic Response}
\begin{tabularx}{\textwidth}{|XXX|}
\toprule
\textbf{User Input} & \textbf{Next Question} & \textbf{Empathetic Response} \\
\midrule
\hline
"I had a medical emergency last week and need to file a claim." &
"Can you please provide details about the incident?" &
"I understand that medical emergencies can be stressful. Please share the incident details so we can assist you." \\
\midrule
\hline
"I've received a medical bill that seems incorrect." &
" Could you please provide me with the details of the bill?" &
"Dealing with medical bills can be confusing. Please share the bill details, and we'll look into it." \\
\midrule
\hline
"I'm getting a '404 Not Found' error on my website. What should I do?" &
"Can you provide more context about when and where it occurs?" &
"Website errors can be frustrating. Please tell me more, so I can assist you effectively." \\
\midrule
\hline
"My code won't compile, and I don't understand the error message." &
"Could you share the error message and a snippet of your code for better assistance?" &
"Coding errors can be challenging. Please provide the error message and your code, and we'll figure it out together." \\
\bottomrule

\end{tabularx}
\caption{Generated examples for training a question generation model with empathy \label{empathy}}
\end{table}

\clearpage

\section{HR Use Case}
\label{usecase}
HR-Agent systems are used to request leaves, inquire about benefits, retrieve payroll details, apply for internal jobs, navigate the onboarding process, schedule performance reviews, request training, report workplace issues, access policies, participate in surveys, and engage with HR initiatives. It simplifies benefit enrollment, goal setting, safety guidelines, and compliance training.

\section{Data Generation Process}
While there are many Schema-Guided Dialogue datasets discussed in the previous section, these datasets are not HR-specific. Therefore, we need to generate synthetic data for creating a TOD system. We employ Claude for generating synthetic data, which is more cost-effective than \textsc{GPT-4} in terms of API costs and easily integrates with the AWS Ecosystem. In our search for a cost-effective data generation option, we attempted to use Claude to generate both conversations and schemas. However, after trying various prompts, we found that the generated schema does not always contain entities from the conversation (non-extractive), fails to capture correct information, and remains inconsistent. The chatbot's questions can also become redundant. We provide an example and discuss other options and problems in Appendix~\ref{claude_schema}. These issues collectively make the synthetic data challenging to use.

As mentioned in the previous section, most models have performed two tasks. The first task is to select the relevant entity, and the second task is to identify the correct answer for that entity~\citep{zhang-etal-2020-find}. Therefore, we considered generating data that can be used for both of these cases and using separate models for each case. We transformed the synthetic data into a different format, where we asked the model to provide the following: utterance, some questions from the same domain, relevant questions that could be asked based on the utterance, and the extracted entity for each answer.

We follow~\citep{taori2023alpaca, wang2023selfinstruct, gunasekar2023textbooks} to create prompts. We use a batch process by explicitly asking the model to produce multiple samples. We increase the diversity of papers by randomly selecting a domain from the list of the following domains (See Appendix~\ref{usecase} for all HR use cases), choosing a random number of questions and a random number of answers. Thus, our proposed model can cover all these use cases and are more versatile.
We also provide a few examples and randomly select one to put into prompt for each generation. Our example prompt is shared in Table~\ref{prompt_claude}, and we present an example of generated data in Table~\ref{example_calude}. We have also tried other model options but none of them work as illustrated in Appendix~\ref{other_option} 

\end{document}